\documentclass{article}

\usepackage{microtype}
\usepackage{graphicx}
\usepackage{subcaption}
\usepackage{booktabs} 
\usepackage{multirow}

\usepackage{hyperref}

\usepackage[preprint]{icml2026}

\usepackage{amsmath}
\usepackage{amssymb}
\usepackage{mathtools}
\usepackage{amsthm}
\usepackage{xcolor}
\usepackage[table]{xcolor}

\usepackage[capitalize,noabbrev]{cleveref}

\theoremstyle{plain}

\theoremstyle{definition}

\theoremstyle{remark}

\usepackage[textsize=tiny]{todonotes}

\icmltitlerunning{Rethinking Practical and Efficient Quantization Calibration for Vision-Language Models}
\begin{document}

\twocolumn[
  \icmltitle{Rethinking Practical and Efficient Quantization Calibration for Vision-Language Models}



  \icmlsetsymbol{equal}{*}

  \begin{icmlauthorlist}
    \icmlauthor{Zhenhao Shang}{equal,npu}
    \icmlauthor{Haizhao Jing}{equal,npu}
    \icmlauthor{Guoting Wei}{nj}
    \icmlauthor{Haokui Zhang}{npu}
    \icmlauthor{Rong Xiao}{intellif}
    \icmlauthor{Jianqing Gao}{iflytek}
    \icmlauthor{Peng Wang}{npu}

  \end{icmlauthorlist}

  \icmlaffiliation{npu}{Northwestern Polytechnical University, Xi'an, China}
  \icmlaffiliation{nj}{Nanjing University of Science and Technology, Nanjing, China}
  \icmlaffiliation{intellif}{Intellifusion, Shenzhen, China}
  \icmlaffiliation{iflytek}{iFLYTEK, China}

  \icmlcorrespondingauthor{Haokui Zhang}{hkzhang@nwpu.edu.cn}
  \icmlcorrespondingauthor{Peng Wang}{peng.wang@nwpu.edu.cn}

  \icmlkeywords{Machine Learning, ICML}

  \vskip 0.3in
]



\printAffiliationsAndNotice{}  

\begin{abstract}
Post-training quantization (PTQ) is a primary approach for deploying large language models without fine-tuning, and the quantized performance is often strongly affected by the calibration in PTQ. By contrast, in vision–language models (VLMs), substantial differences between visual and text tokens in their activation distributions and sensitivities to quantization error pose significant challenges for effective calibration during PTQ.
In this work, we rethink what PTQ calibration should align with in VLMs and propose the \textbf{T}oken-level Importance-aware \textbf{L}ayer-wise \textbf{Q}uantization framework (TLQ). Guided by gradient information, we design a token-level importance integration mechanism for quantization error, and use it to construct a token-level calibration set, enabling a more fine-grained calibration strategy. Furthermore, TLQ introduces a multi-GPU, quantization-exposed layer-wise calibration scheme. This scheme keeps the layer-wise calibration procedure consistent with the true quantized inference path and distributes the complex layer-wise calibration workload across multiple RTX3090 GPUs, thereby reducing reliance on the large memory of A100 GPUs. TLQ is evaluated across two models, three model scales, and two quantization settings, consistently achieving performance improvements across all settings, indicating its strong quantization stability. The code will be released publicly.

\end{abstract}

\section{Introduction}

\begin{figure}[h]
    \centerline{\includegraphics[width=\columnwidth]{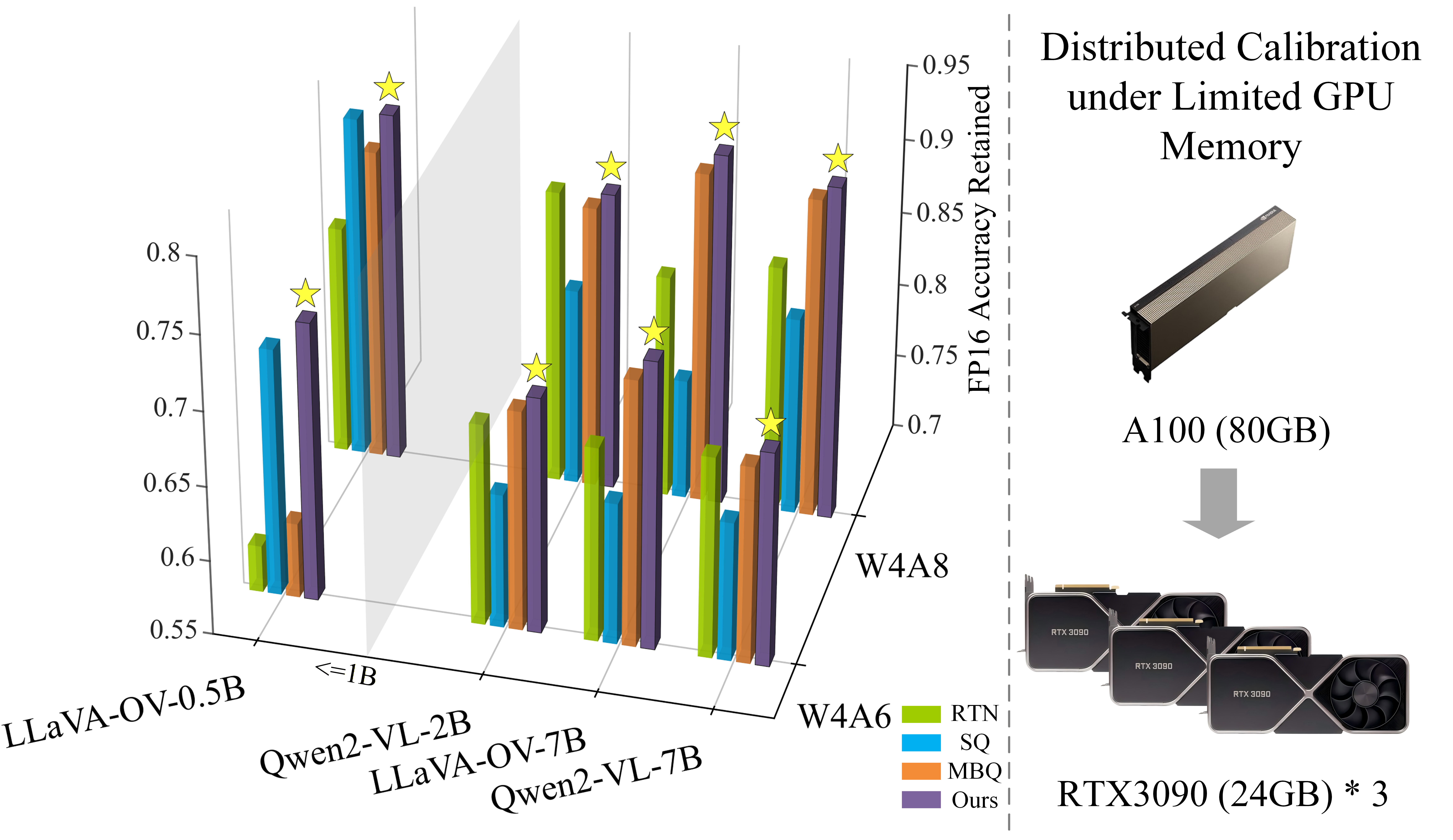}}
    \caption{\textbf{Left}: TLQ is compared with three representative PTQ baselines across two vision–language models with three model scales under two quantization settings. TLQ achieves the highest percentage of FP16 accuracy retained after quantization across all settings, demonstrating strong stability. \textbf{Right}: Illustration of the hardware efficiency of the proposed framework, which reduces the requirement from a single 80GB A100 GPU to a practical setup using three 24GB RTX3090 GPUs.}
    \label{img：visual}
\end{figure}

Large-scale vision-language models (VLMs) have recently demonstrated remarkable capabilities across a wide range of multimodal understanding and real-world tasks, including visual question answering, image captioning, and embodied reasoning, by jointly modeling visual and textual modalities at scale \cite{alayrac2022flamingo,li2023blip,liu2023visual}. Despite their strong performance, deploying such models in practical scenarios remains prohibitively expensive, particularly during inference, due to their massive parameter sizes and intensive compute and memory demands. Among existing compression techniques, post-training quantization (PTQ) has emerged as a particularly attractive solution, as it enables low-bit inference without retraining or fine-tuning, offering favorable trade-offs in efficiency, cost, and deployment simplicity \cite{frantar2022gptq,xiao2023smoothquant}. 

While PTQ has been extensively studied for large language models (LLMs), existing calibration strategies are not fully suited to VLMs, as they largely overlook the discrepancy between token statistics and token importance induced by multimodal token structures. In VLMs, visual tokens not only differ substantially from textual tokens in activation distributions and sensitivities to quantization error, but also dominate the token sequence by a large margin while being highly redundant due to spatial correlations in visual representations \cite{dosovitskiy2020image,bolya2022token,xu2022evo}. As a result, calibration objectives that aggregate token-level statistics are disproportionately influenced by a large number of low quantization importance tokens, leading to systematically biased scale estimation and more severe calibration errors than those observed in large language models.

In contrast, most existing PTQ methods perform scale search by global statistics \cite{lin2024awq,williams2024impact}, implicitly assuming that all tokens contribute equally to calibration quality. Motivated by the resulting modality imbalance, several recent VLM-oriented PTQ approaches introduce modality-level reweighting schemes that assign fixed or learnable coefficients to visual and textual tokens during calibration \cite{wang2024q,li2025mbq,yu2025mquant}. However, these methods remain insufficient for fully addressing the calibration challenges in VLMs, as they operate at a coarse modality level and do not explicitly account for token-level variations in sensitivity to quantization error. This observation highlights a challenge in VLM-PTQ calibration strategies: by ignoring token-level importance for quantization error, the calibration objective remains misaligned with inference-critical tokens.

 In light of this challenge, token-level importance with respect to quantization error needs to be explicitly characterized in VLM post-training quantization. At a fundamental level, the objective of PTQ is to preserve the model’s inference behavior after quantization by minimizing deviations in task-relevant outputs. From this perspective, an effective calibration process should prioritize tokens whose perturbations exert a larger impact on the model’s final outputs, rather than treating all tokens as equally important. In a training-free setting, gradient signals naturally provide a principled measure of token importance, as they directly quantify the sensitivity of the inference loss with respect to token-level perturbations \cite{pmlr-v119-nagel20a,pmlr-v139-hubara21a}. Importantly, these gradients can be obtained without any parameter updates, making them well aligned with the constraints of PTQ.

Building on these observations, we refine the PTQ paradigm for VLMs and propose a training-free Token-level Importance-aware Layer-wise Quantization (TLQ) framework. Specifically, we leverage gradient information to characterize token-wise sensitivity to quantization perturbations and construct a compact yet information-dense subset of tokens for quantization hyperparameters estimation. By operating at the token level, our approach enables fine-grained calibration that naturally balances visual and textual tokens, in contrast to prior methods that rely on coarse modality-level weighting schemes learned or heuristically tuned during calibration.
Furthermore, to improve quantization accuracy while maintaining practical feasibility, TLQ introduces a multi-GPU, quantization-exposed layer-wise calibration strategy. This strategy aligns calibration with the true quantized inference path by propagating reweighted inference information across layers. Meanwhile, it splits the quantization-related computation into three parts and distributes them across three GPUs, communicating only the necessary intermediate statistics required for scale estimation. As a result, this design substantially alleviates the memory bottleneck inherent to complex layer-wise PTQ, enabling a quantization task that would otherwise require a single A100 to run on three RTX3090 GPUs. As illustrated in \cref{img：visual}, extensive experiments conducted across three different model scales of two representative VLMs (LLaVA-onevision \& Qwen2-VL), under two quantization settings (W4A6 \& W4A8), demonstrate that TLQ consistently delivers strong performance improvements and maintains stable behavior across different quantization settings.

In summary, this work makes the following contributions:
\begin{itemize}
  \setlength{\itemsep}{2pt}
  \setlength{\parskip}{2pt}
    \item We rethink the calibration objective of post-training quantization in VLMs and identify token-level misalignment as a key source of calibration bias. To address this, we use gradients as guidance and design a token-level importance integration mechanism for quantization error. This mechanism accumulates gradient-based importance estimates for each token and selects the most important tokens to construct a token-level calibration set, thereby refining calibration at a finer granularity.
    \item We introduce a layer-wise, quantization-exposed distributed calibration strategy. By explicitly aligning calibration with the true quantized inference path, we correspondingly decompose the quantization-related computations and dispatch them across multiple GPUs, while communicating only the calibration statistics among GPUs. This enables complex post-training quantization of VLMs to be performed on multiple RTX3090 GPUs, rather than being restricted to a single large-memory GPU such as A100.
    \item The experimental results show that TLQ consistently achieves performance improvements over three strong baseline methods across three model scales of two VLMs under two quantization settings. This not only demonstrates the effectiveness of TLQ, but also indicates its strong quantization stability.
\end{itemize}

\section{Related Work and Preliminaries}
\subsection{PTQ for VLMs}
PTQ has become a practical solution for deploying large transformer models by reducing memory footprint and inference cost without additional fine-tuning. Typical PTQ pipelines determine quantization granularity, clipping ranges, and scaling factors for weights and activations using a small calibration set, where calibration parameters are optimized by minimizing reconstruction error under quantized forward passes. In large language models (LLMs), extensive prior work has demonstrated that carefully designed, training-free calibration can preserve accuracy even at low bit-widths \cite{dettmers2022gpt3,yuan2023rptq,dettmers2023spqr,shao2023omniquant}. These methods largely assume unimodal token distributions and token-agnostic calibration objectives, and they now form the foundation of many quantization pipelines.

However, directly applying the aforementioned unimodal PTQ strategies to VLMs is highly challenging. VLMs jointly process visual and textual tokens, which differ substantially in activation scale and sensitivity to quantization error. Moreover, visual tokens typically dominate the token sequence in quantity and exhibit substantial spatial redundancy,
a phenomenon widely observed in vision transformers and token pruning studies\cite{bolya2022token,rao2021dynamicvit,yin2022vit}. As a result, during quantization calibration, tokens that are numerically dominant but less sensitive to quantization error tend to exert disproportionate influence on the calibration process, inducing calibration bias and ultimately impairing the accuracy of quantization.

Recent work has begun to explicitly address these challenges by incorporating modality-aware or token-aware calibration mechanisms. Q-VLM \cite{wang2024q} investigates post-training quantization for VLMs and shows that naïve layer-wise independent calibration is insufficient, proposing a strategy that accounts for cross-layer dependency during calibration. MBQ \cite{li2025mbq} further observes that visual and textual tokens exhibit different loss sensitivities and introduces modality-balanced calibration, where gradient-based signals are used to reweight visual and textual contributions during scale search. VLMQ \cite{xue2025vlmq} approaches the problem from a token-importance perspective, augmenting Hessian-based PTQ with token-level importance factors to suppress redundant visual tokens while preserving salient ones. While these methods introduce token- or modality-aware signals into calibration, they still rely on aggregated or coarse-grained objectives during scale search, leaving token-level alignment with quantization error only partially addressed.

Taken together, these works reveal a consistent trend that the primary limitation of existing VLM post-training quantization methods resides in the calibration stage, where calibration objectives fail to faithfully capture token-level sensitivity to quantization error under heterogeneous modality distributions. This understanding further motivates the exploration of more fine-grained, quantization-error-sensitivity-aware calibration paradigms, which motivates a rethinking of PTQ calibration objectives toward explicit token-level alignment with quantization error, as explored in this work.

\begin{figure*}[t]
\centering
\includegraphics[width=\linewidth]{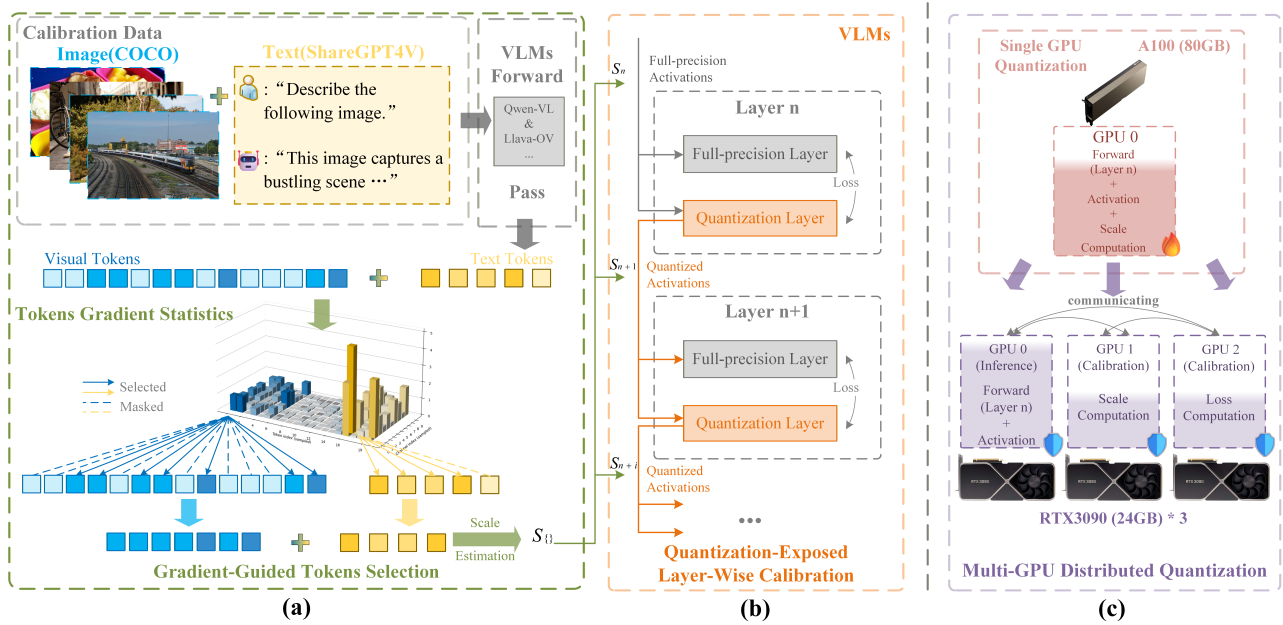}
\caption{Overview of our proposed \textbf{T}oken-level Importance-aware \textbf{L}ayer-wise \textbf{Q}uantization framework. \textbf{(a)} Gradient-guided token selection identifies output-sensitive tokens for calibration, reducing token-agnostic bias from redundant tokens.
\textbf{(b)} Quantization-exposed layer-wise calibration propagates quantized activations across layers to align calibration with the inference path. \textbf{(c)} Memory-efficient calibration under deployment constraints is achieved by decoupling layer-wise inference, scale computation, and loss evaluation across multiple GPUs, enabling calibration on three 24GB RTX3090 GPUs instead of a single 80GB A100.}
\label{fig:methods}
\end{figure*}

\subsection{Quantization Calibration}
\textbf{The presence of outliers significantly increases the quantization error.}
For an FP16 tensor, including both weights and activations, the quantization to low-bit integer values can be summarized by the following formulation:
\begin{equation}
    \hat{X}_{i, j} = \mathrm{round}(\frac{X_{i, j}}{s_i}),
\end{equation}
\begin{equation}
    s_i = \frac{\max_j|X_{i, j}|}{2^{n-1}-1}
\end{equation}
For simplicity, we focus on symmetric quantization here; the discussion for asymmetric quantization follows in a similar manner. Here, $X$ denotes an FP16-precision tensor, $\hat{X}$ denotes the quantized low-bit integer, and n represents the quantization bit width. When $X = A \in \mathbb{R}^{N \times C}$, it corresponds to per-token quantization of activations, and when $X = W \in \mathbb{R}^{C_1 \times C_2}$, it corresponds to per-channel quantization of weights.

For a given quantization bit width $n$, the range of the quantized integers is $[-2^{n-1}, 2^{n-1}-1]$. For example, when quantizing to int8, the range is $[-128, 127]$. The corresponding quantization step size $\Delta$ is:
\begin{equation}
    \Delta = \frac{2\max_j|X_{i, j}|}{2^n-1}
\end{equation}
Meaning that all floating-point values within each interval of width are mapped to the same integer after quantization.
In theory, under round-to-nearest quantization, the rounding error $e$ follows a uniform distribution: 
\begin{equation}
    e \sim \mathcal{U}(-\frac{\Delta}{2}, \frac{\Delta}{2})
\end{equation}
Its mean is zero and its variance is $\frac{\Delta^2}{12}$.
The expected error energy (i.e., the expected squared L2 norm) can therefore be expressed accordingly:
\begin{equation}
    \mathrm{E}[\lVert e \rVert_2^2] = \frac{\Delta^2}{12}
\end{equation}
Indicating that the quantization error is positively correlated with the square of the step size. Consequently, to reduce quantization error, it is necessary to decrease the quantization step size $\Delta$. However, the widespread presence of outliers in large models significantly increases the quantization step size, leading to substantial quantization errors.

\textbf{Introduction to previous methods that reduce quantization error through calibration.}
To reduce the error of activation quantization, previous works have proposed various calibration methods. Among them, per-channel smoothing is a classical and widely used approach. It works by dividing the input activations by per-channel scaling factors  and multiply it to the corresponding linear layer weights. 
For a given linear layer with input activations $X \in \mathbb{R}^{N \times C_1}$ and weights $W \in \mathbb{R}^{C_2 \times C_1}$, the output $Y \in \mathbb{R}^{N \times C_2} $ of the linear layer can be expressed as follows:
\begin{equation}
    Y = (X\mathrm{diag}(scale)^{-1})(\mathrm{diag}(scale)W) = \tilde{X}\tilde{W}
\end{equation}
Here, $scale \in \mathbb{R}^{C_1}$ is a per-channel scaling factor, it is defined as follows:
\begin{equation}
    scale = \sqrt{\max|X|/\max|W|}
\end{equation}
In general, outliers in activations are more pronounced than those in weights. This method shifts the activation outliers to the weights, significantly reducing 
$ \max|X|$ and thus the quantization step size $\Delta$, while preventing 
$ \max|W|$ from becoming too large, thereby reducing the overall quantization error.
Moreover, the operation of dividing activations by the scale can be fused into the preceding layer, such as RMSNorm or another linear layer, so that the model does not incur extra computation during inference.

\section{Method}
\label{section:method}



In this section, we first provide an in-depth analysis and explanation of how quantization errors affect the final outputs, and then present our \textbf{T}oken-level Importance-aware \textbf{L}ayer-wise \textbf{Q}uantization framework (TLQ). As shown in \cref{fig:methods}, TLQ consists of Token-level Importance-Aware Quantization Calibration and Joint Optimization of Layer-wise Quantization and Memory Footprint.

\subsection{Token-Level Importance Sensitivity for Quantization Calibration}
\subsubsection{Tokens Importance}
Since quantization error is inevitable, our primary objective should be to minimize the error in the model’s final output. Some previous studies have primarily focused on quantization errors introduced by outliers in the model. However, the impact of quantization errors varies across different tokens, and their influence on the final output can be characterized by the gradients with respect to the model output. Tokens with extremely small gradients have a negligible effect on the final result even if they incur quantization errors. Under fixed weights, the relationship between the model’s final output $Y$, its quantization error $\Delta Y$, activation quantization error of a certain layer $\delta x_i$, and the corresponding gradient $g_i$ can be expressed by the following equation.
\begin{equation}
    \Delta Y \approx \sum_i^N g_i^T \delta x_i
\end{equation}
\begin{equation}
    \delta x_i = Q(x_i) - x_i
\end{equation}

This phenomenon is more pronounced in VLMs, where a large number of visual tokens introduce substantial redundancy. As a result, many visual tokens exhibit gradients close to zero, leading to a significantly enlarged gradient disparity across tokens. Naturally, treating all tokens equally is suboptimal. Instead, we focus more on tokens with large gradients and reduce their quantization errors, which is crucial for minimizing the model's final output quantization error.

\subsubsection{Gradient-Guided Tokens Selection}
For activation quantization, the calibration-time smoothing coefficient $scale$ is per-channel so that it can be fused into the preceding layer, the quantization-time scaling factor $s_i$ is per-token to account for differences in activation distributions across tokens. This design leads to the tokens closest to the $scale$ have the smallest quantization step sizes $\Delta$ and therefore the lowest quantization errors. Under this setting, conventional methods that treat all tokens equally may be biased by visual tokens with large magnitudes but near-zero gradients when calculating $scale$, leading to suboptimal performance.

To ensure that important tokens have smaller quantization errors $\delta x_i$, their quantization step sizes $\Delta$ should be smaller, that is, the difference between $\max|x_i|$ and $\min|x_i|$ across all channels of the important token should be minimized. A naive approach would be to set $scale$ equal to the token value itself, which would make the quantization step size $\Delta$ zero for that token after smoothing($x/scale$). However, for each input of $N$ tokens, a subset of $K$ tokens is identified as important(shape $K \times C$, with $1<K<N$), while the scale is a single vector of shape $C$. Therefore, we divide all input tokens into two groups: important and non-important, and compute the scale based only on the important tokens as follows.
\begin{equation}
\label{equation:scale}
    scale = x\_stat_c^r
\end{equation}
\begin{equation}
    x\_stat_c = \max_{i \in I}|X_{i, c}|, c=1, 2,...,C
\end{equation}
Here $I$ denotes the important selected tokens with the highest gradients. $r$ denotes the $ratio$ obtained through a search procedure, which is used to balance the overall magnitude of the $scale$. A detailed description is provided in Section \ref{section: per_layer quantization}.

During calibration, we need to select important tokens from the B input samples, where B=128 in our setting. However, the indices of important tokens vary across different samples, and the distribution of important tokens in the calibration set also differs from that in real inference data. Therefore, we aim to adopt a selection strategy that enables the identified important tokens to better match the real inference data distribution.
Here we adopt a simple yet effective Top-K token aggregation strategy as illustrated in figure \ref{fig:methods} (a), which explicitly highlights the conceptual contribution of computing the scaling factor based on the most important tokens.

Specifically, from a global perspective, we first compute gradients for each of the $B$ input samples, then sum the absolute values of these gradients and take the mean along the channel dimension $C$, resulting in an overall gradient statistic $sum_n$ of shape $N$.
\begin{equation}
    sum_n = \sum_{b=1}^B \lVert g_{b, n}\rVert, n=1,...,N
\end{equation}
Based on this statistic, we select the top 50\% tokens as important tokens, and use them to compute the smoothing coefficient $scale$ for activation quantization.
\begin{equation}
    I = \mathrm{Top}_{50\%}(\{sum_n|n=1,...,N\})
\end{equation}

\subsection{Quantization Calibration under Deployment Constraints}
\subsubsection{Quantization-Exposed Layer-Wise Calibration}
\label{section: per_layer quantization}

Current methods not only compute $x\_stat$ from input activations, but also search for an optimal $ratio$ based on the quantization loss to obtain the $scale$. However, for large models, searching over the entire network is impractical. As a result, a layer-wise search strategy with lower memory overhead is typically adopted. The core idea is to search an optimal $ratio$ to minimize the quantization error of each layer individually. Its search process can be formulated as follows.
\begin{equation}
    scale^{(l)} = (x\_stat^{(l)})^{r^{(l)}}
\end{equation}
\begin{equation}
    r^{(l)*} = \arg \min_r \mathbb{E} \left[\lVert f_{fp}^{(l)}(x^{(l)})-f_q^{(l)}(x^{(l)};scale^{(l)}) \rVert_2^2\right]
\end{equation}
Here, $l$ denotes the current layer being searched, $r$ denotes the $ratio$ to be optimized, and $x^{(l)}$ denotes the input to the current layer. $f_q^{(l)}(\cdot)$ denotes the quantized function, $f_{fp}^{(l)}(\cdot)$ denotes the original full-precision function. The corresponding computation formulas are given as follows:
\begin{equation}
    x^{(l)} = f_{fp}^{(l-1)}(x^{(l-1)})
\end{equation}
\begin{equation}
f_{fp}^{(l)}(x^{(l)}) = W^{(l)} \cdot x^{(l)} + b^{(l)}
\end{equation}
\begin{equation}
\begin{aligned}
f_q^{(l)}(x^{(l)};scale^{(l)})
=  Q_w(W^{(l)}\cdot scale^{(l)}) \\
\cdot Q_a\!\left(\frac{x^{(l)}}{scale^{(l)}}\right) + b^{(l)}
\end{aligned}
\end{equation}

The limitation of this approach is that there exists a gap between minimizing per-layer error and minimizing the final output error of the model.
We aim to address this gap by enabling each layer, during its search, to be aware of errors introduced by other layers, while keeping the practical layer-wise search strategy intact. Leveraging the fact that, during model inference, data flows unidirectionally from shallow to deep layers, we propose a quantized activation propagation strategy, allowing deeper layers to perceive errors accumulated from earlier layers. Specifically, we propose two implementations. 

\begin{itemize}
    \item PassAct1: Maintaining two separate streams of input data for each layer, one with full-precision and one with the quantized activations from the previous layer. The scale is searched on the quantized input such that its difference from the full-precision output is minimized.
    \begin{equation}
        r^{(l)*} = \arg \min_r \mathbb{E} \left[\lVert f_{fp}^{(l)}(x_{fp}^{(l)})-f_q^{(l)}(x_q^{(l)};scale^{(l)}) \rVert_2^2\right]
    \end{equation}
    \begin{equation}
        x_{fp}^{(l)} = f_{fp}^{(l-1)}(x_{fp}^{(l-1)})
    \end{equation}
    \begin{equation}
        x_q^{(l)} = f_q^{(l-1)}(x_q^{(l-1)};scale^{(l-1)})
    \end{equation}
    \item PassAct2: As illustrated in figure \ref{fig:methods} (b), replace the full-precision input activations of each layer with the quantized outputs of the previous layer, i.e.:
    \begin{equation}
        x^{(l)} = f_q^{(l-1)}(x^{(l-1)};scale^{(l-1)})
    \end{equation}
\end{itemize}

Comparing the two methods, The first method is more fine-grained, as each layer can fully perceive the quantization errors accumulated from all preceding layers. However, given that the optimizable variable $ratio$, only affects the $scale$ whose shape is $C$, it is intuitively difficult to effectively reduce errors along the token dimension. This may lead to excessively large, uncorrectable errors accumulated from previous layers, which in turn adversely affect the quantization error of the current layer. The second one is more straightforward, essentially making the input of each layer better match the true data distribution. Therefore, we ultimately adopt the second method, and our experiments show that it performs better.

\subsubsection{Distributed Calibration under Limited GPU Memory}
Even with layer-wise calibration, memory consumption remains a severe challenge. During calibration, it is not only required to perform inference on the current layer; we also need to compute the $scale$, search over the $ratio$, calculate the $loss$. This process requires storing multiple intermediate values with shapes equal to the input activations, which can lead to catastrophic memory explosion. The memory consumption of the original layer-wise calibration method can be expressed as follows, here $\|\cdot\|$ denotes the memory footprint of a tensor.
\begin{equation}
\begin{aligned}
\mathcal{M}_{\mathrm{baseline}}^{(l)}
\;\propto\;&
\underbrace{\|\mathbf{x}^{(l)}\|}_{\text{input}}
+
\underbrace{\|\mathbf{y}_{\mathrm{fp}}^{(l)}\|}_{\text{FP output}}
+
\underbrace{\|\mathbf{y}_{\mathrm{q}}^{(l)}(r)\|}_{\text{Q output}}\\
&
+
\mathcal{M}_{\mathrm{layer}}^{(l)}
+
\mathcal{O}\!\left(\|\mathbf{x}^{(l)}\|\right)
\end{aligned}
\end{equation}



Based on the layer-wise calibration framework, we propose an inter-GPU communication strategy that decouples inference, scale computation, and loss computation across different GPUs. The GPUs used for inference are determined by the device map during model loading, while scale computation and loss evaluation are performed on the GPU with the lowest current memory usage. 
After the inference GPU produces the outputs, they are transferred to the GPU responsible for loss computation, so that the peak memory usage on the inference GPU only includes the input tensors, the current layer, and a single output tensor.
\begin{equation}
\mathcal{M}_{g_{\mathrm{infer}}}^{(l)}
\;\approx\;
\mathcal{M}_{\mathrm{layer}}^{(l)}
+
\|\mathbf{x}^{(l)}\|
+
\|\mathbf{y}^{(l)}\|
,
\quad
\mathbf{y}^{(l)} \in
\{\mathbf{y}_{\mathrm{fp}}^{(l)},\mathbf{y}_{\mathrm{q}}^{(l)}\}
\end{equation}
\begin{equation}
\mathcal{M}_{g_{\mathrm{cal}}}^{(l)}
\;\approx\;
\|\mathbf{y}_{\mathrm{fp}}^{(l)}\|
+
\|\mathbf{y}_{\mathrm{q}}^{(l)}(r)\|
+
\mathcal{O}(C)
\end{equation}
By distributing inference and loss calculation across different GPUs, the peak memory consumption is reduced from a summation of all intermediate tensors to the maximum memory usage of a single GPU. As a result, a 7b model calibration process that originally required a single 80GB A100 GPU for a calibration set of 128 samples can now be executed using only three 24GB RTX3090 GPUs, significantly reducing GPU memory consumption during calibration, making our method accessible on widely available hardware.

\begin{table*}[h]
\centering
\footnotesize
\renewcommand{\arraystretch}{0.9}
\caption{Main results of LLaVA-onevision-7b, Qwen2-VL-7b. SQ is short for smoothquant. To ensure fair comparison and consistency with prior evaluations, we reuse the results of overlapping experimental settings reported in MBQ. Methods marked with * means the results of the first four benchmarks are reused.}
\label{tab:main}
\resizebox{\textwidth}{!}{
\begin{tabular}{lccccccccc}
\toprule
Model & Bitwidth & Method & MMMU & OCRBench & VizWiz & TextVQA & ChartQA & SEEDBench2plus & Average($\uparrow$) \\
\midrule
\multirow{9}{*}{LLaVA-onevision-0.5b} & fp16 & - 
& 32.4 & 57.6 & 46.6 & 65.8 & 61.4 & 52.9 & 52.8 \\
\cmidrule{2-10}
&\multirow{4}{*}{W4A6}
&RTN  & 26.3 & 33.8 & 15.6 & 42.6 & 37.0 & 29.0 & 30.7 \\
&&SQ   & 27.2 & 39.9 & \textbf{36.8} & 45.7 & 39.7 & \textbf{37.3} & 37.8 \\
&&MBQ  & 28.0 & 37.6 & 21.8 & 43.9 & 39.4 & 19.4 & 31.7 \\
&&TLQ & \textbf{28.1} & \textbf{42.9} & 32.2 & \textbf{49.7} & \textbf{45.4} & 35.3 & \textbf{38.9} \\

\cmidrule{2-10}
&\multirow{4}{*}{W4A8}
&RTN  & 28.1 & 44.2 & 18.8 & 50.8 & 44.6 & 36.8 & 37.2 \\
&&SQ   & 26.3 & 44.3 & \textbf{38.8} & 51.9 & 44.8 & \textbf{41.2} & 41.2 \\
&&MBQ  & 26.8 & 45.9 & 26.4 & 55.3 & 48.6 & 35.6 & 39.8 \\
&&TLQ & \textbf{29.2} & \textbf{51.0} & 33.6 & \textbf{55.5} & \textbf{49.3} & 36.2 & \textbf{42.5} \\

\midrule
\multirow{9}{*}{LLaVA-onevision-7b} & fp16 & - 
& 46.0 & 62.2 & 60.4 & 76.1 & 80.0 & 64.8 & 64.9 \\
\cmidrule{2-10}
&\multirow{4}{*}{W4A6} & 
RTN  & 37.9 & 42.3 & 53.8 & 59.5 & 69.6 & 60.4 & 53.9 \\
&&SQ   & 38.3 & 35.9 & 53.7 & 56.5 & 64.9 & 60.0 & 51.6 \\
&&MBQ  & 40.6 & 48.1 & \textbf{56.3} & 65.4 & 71.5 & 60.0 & 57.0 \\
&&TLQ & \textbf{41.4} & \textbf{49.8} & \textbf{56.3} & \textbf{65.9} & \textbf{73.0} & \textbf{60.7} & \textbf{57.9} \\

\cmidrule{2-10}
&\multirow{4}{*}{W4A8} & 
RTN*  & 38.2 & 40.1 & 58.2 & 61.5 & 70.6 & 62.4 & 55.2 \\
&&SQ*   & 30.9 & 32.0 & 56.7 & 56.9 & 66.4 & 61.4 & 50.7 \\
&&MBQ*  & 42.6 & 52.3 & \textbf{58.9} & 68.3 & 74.9 & 62.4 & 59.9 \\
&&TLQ & \textbf{45.9} & \textbf{53.3} & 58.7 & \textbf{68.8} & \textbf{75.3} & \textbf{62.8} & \textbf{60.8} \\

\midrule
\multirow{9}{*}{Qwen2-VL-2b} & fp16 & -
& 40.8 & 76.5 & 66.0 & 79.5 & 72.3 & 62.3 & 66.2 \\
\cmidrule{2-10}
&\multirow{4}{*}{W4A6}
&RTN  & 33.7 & 61.0 & 55.9 & 69.6 & 58.0 & 53.7 & 55.3 \\
&&SQ   & 33.9 & 59.4 & 55.2 & 66.8 & 46.9 & 51.5 & 52.3 \\
&&MBQ  & \textbf{34.6} & 59.8 & 55.2 & 70.1 & 61.9 & 54.8 & 56.1 \\
&&TLQ & 33.8 & \textbf{63.5} & \textbf{56.9} & \textbf{71.8} & \textbf{62.4} & \textbf{54.9} & \textbf{57.2} \\
\cmidrule{2-10}
&\multirow{4}{*}{W4A8}
&RTN  & 35.4 & 66.2 & 57.9 & \textbf{74.9} & 65.0 & 57.4 & 59.5 \\
&&SQ   & 33.2 & 57.6 & 56.5 & 70.4 & 58.9 & 52.8 & 54.9 \\
&&MBQ  & 34.8 & 65.4 & 56.4 & 73.4 & 63.9 & 57.3 & 58.5 \\
&&TLQ & \textbf{36.3} & \textbf{66.7} & \textbf{58.5} & 74.1 & \textbf{65.4} & \textbf{57.7} & \textbf{59.8} \\

\midrule
\multirow{9}{*}{Qwen2-VL-7b} & fp16 & - 
& 50.6 & 80.7 & 68.3 & 82.0 & 81.4 & 69.4 & 72.1 \\

\cmidrule{2-10}
&\multirow{4}{*}{W4A6}
&RTN  & 40.8 & 62.0 & \textbf{57.0} & 68.1 & 70.6 & 62.7 & 60.2 \\
&&SQ   & 40.2 & 55.9 & 51.4 & 64.1 & 68.4 & 63.2 & 57.2 \\
&&MBQ  & 41.1 & 65.8 & 49.8 & \textbf{70.0} & 70.8 & 62.3 & 60.0 \\
&&TLQ & \textbf{41.9} & \textbf{68.4} & 52.9 & 69.4 & \textbf{72.2} & \textbf{64.3} & \textbf{61.5} \\

\cmidrule{2-10}
&\multirow{4}{*}{W4A8}
&RTN*  & 43.8 & 60.3 & 58.9 & 71.0 & 74.5 & \textbf{66.5} & 62.5 \\
&&SQ*   & 45.9 & 57.1 & 52.3 & 68.2 & 71.6 & 65.3 & 60.1 \\
&&MBQ*  & \textbf{47.2} & 72.8 & 59.3 & 75.0 & 75.7 & \textbf{66.5} & 66.1 \\
&&TLQ & 45.9 & \textbf{74.2} & \textbf{61.2} & \textbf{75.8} & \textbf{77.1} & 66.3 & \textbf{66.8} \\

\bottomrule
\end{tabular}
}
\end{table*}

\section{Experiments}

\subsection{Experimental Setup}

\textbf{Calibration Datasets.} 
In our experiments, we construct the calibration set using the enhanced COCO Caption dataset released by ShareGPT4V \cite{chen2024sharegpt4v}. Following prior practice, we randomly sample 128 image–caption pairs for calibration. Each caption in this dataset is generated by GPT-4V, providing high-quality and semantically rich descriptions aligned with the visual content. For each evaluated VLM, the sampled image–caption pairs are further formatted using the model-specific conversation templates, yielding instruction-style inputs that are consistent with the model’s inference protocol.

\textbf{Evaluation Datasets.} 
To comprehensively evaluate the performance of our proposed TLQ across diverse vision–language understanding scenarios, we adopt multiple widely used multimodal benchmarks, including MMMU-Val \cite{yue2024mmmu}, OCRBench \cite{liu2024ocrbench}, VizWiz-VQA-Val \cite{gurari2018vizwiz}, TextVQA-Val \cite{TextVQA}, ChartQA \cite{masry2022chartqa}, and SeedBench2-Plus \cite{SeedBench2Plus}. These datasets cover a broad range of tasks such as multi-step reasoning across different disciplines, optical character recognition and text understanding in natural images, robustness under real-world visual conditions, visual reasoning over charts, and general multimodal perception and generalization capabilities.

\textbf{Evaluation Models.} 
To validate the effectiveness of our TLQ across VLMs with different architectures and model scales, we conduct experiments on two mainstream VLM architectures, LLaVA-onevision \cite{Llava-onevision} and Qwen2-VL \cite{bai2023qwen}, specifically including LLaVA-onevision-7B, LLaVA-onevision-0.5B, Qwen2-VL-7B, and Qwen2-VL-2B. 
These models differ in architecture and scale, enabling a comprehensive evaluation of our method across diverse VLM designs.

\textbf{Baselines and Quantization Settings.}
To assess the effectiveness and performance gains of our TLQ in activation quantization calibration, we compare against three strong baselines, RTN, SmoothQuant \cite{xiao2023smoothquant}, and MBQ \cite{li2025mbq}, under both W4A8 and W4A6 quantization settings. To ensure a fair comparison of calibration effectiveness, all methods adopt per-token symmetric quantization for activations and per-channel symmetric quantization for weights.

\begin{table*}[t]
\centering
\footnotesize
\renewcommand{\arraystretch}{0.9}
\setlength{\tabcolsep}{12pt}
\caption{Ablation study on LLaVA-onevision-7b with W4A8 quantization on the MMMU \& TextVQA validation datasets. The relationship between x\_stat and the scale is given in the equation \ref{equation:scale}. Here, x\_mean and x\_max are the base methods for computing x\_stat, while we propose the TopK-sum method for its computation.
PassAct denotes quantization-exposed layer-wise calibration.
}
\label{tab:ablation}
\begin{tabular}{c c c c @{\hspace{6pt}\vrule\hspace{6pt}} c c c c c}
\toprule
\multirow{2}{*}{Method} &
\multicolumn{5}{c}{Components} &
\multirow{2}{*}{MMMU($\uparrow$)} &
\multirow{2}{*}{TextVQA($\uparrow$)} \\
\cmidrule{2-4}\cmidrule{5-6}
& x\_mean & x\_max & TopK-Sum & PassAct1 & PassAct2 & & \\
\midrule
base
& \checkmark & - & - & - & - & 43.6 & 67.3 \\
\midrule
\multirow{3}{*}{x\_stat calculation}
& \checkmark & - & - & - & \checkmark & 44.1 & 67.6 \\
& - & \checkmark & - & - & \checkmark & 45.6 & 67.7 \\
& - & - & \checkmark & - & \checkmark & \textbf{45.9} & \textbf{68.8} \\
\midrule
\multirow{3}{*}{PassAct strategy}
& - & - & \checkmark & - & - & 44.1 & 66.9 \\
& - & - & \checkmark & \checkmark & - & 45.7 & 67.7 \\
& - & - & \checkmark & - & \checkmark & \textbf{45.9} & \textbf{68.8} \\
\bottomrule
\end{tabular}
\end{table*}

\subsection{Main Results}

The main experimental results are shown in \cref{tab:main}. From the observations, we can find that:
\begin{itemize}
  \setlength{\itemsep}{2pt}
  \setlength{\parskip}{2pt}
    \item Our proposed TLQ consistently achieves the best average accuracy across all experimental settings. In the LLaVA-onevision-7B model, our method surpasses the second-best MBQ by 0.9\% in average accuracy for both W4A8 and W4A6 settings, and under W4A6, it achieves an accuracy greater than or equal to MBQ in all datasets. In Qwen2VL-7B with W4A6, the average performance is even 1.5\% higher than MBQ.
    \item TLQ demonstrates clear robustness and stability. Indeed, on the LLaVA-onevision-0.5B model, MBQ performs worse than Smoothquant, whereas on the larger 7B model, MBQ consistently outperforms Smoothquant. In contrast, TLQ maintains leading average accuracy across diverse experimental settings, clearly reflecting its stability and robustness.
    \item TLQ achieves significant improvements on smaller models. In theory, smaller models are more sensitive to quantization errors, and MBQ method shows a noticeable drop in accuracy on these models. However, our method refines the calibration process, resulting in substantial accuracy gains over MBQ on small models.
\end{itemize}
\subsection{Ablation Study}



For Token-Level Importance Sensitivity for Quantization Calibration, we conduct an ablation study on several scale computation strategies, including conventional mean-based and max-based methods without token selection, as well as the proposed gradient-based Top-K token selection approach. As shown in \cref{tab:ablation}, in the x\_stat calculation component, when fixing the PassAct strategy to PassAct2, our TopK-Sum method achieves higher accuracy than X\_mean and X\_max on both MMMU and TextVQA, demonstrating the effectiveness of first selecting important tokens based on gradients and then computing the scaling factors.

For Quantization-Exposed Layer-Wise Calibration, we conduct an ablation study by comparing the setting without activation propagation to the two activation propagation strategies introduced in our method.
As shown in the PassAct strategy component of \cref{tab:ablation}, our PassAct2 strategy achieves the best performance on both validation datasets. Moreover, both of our proposed layer-wise activation propagation strategies(PassAct1 \& PassAct2) attain higher accuracy than the conventional approach without activation propagation, further highlighting the importance of considering the overall model error during per-layer quantization.

\subsection{Memory Usage Comparison During Calibration}
To broaden the applicability of our method, we also propose Distributed Calibration under Limited GPU Memory, which significantly reduces single-GPU peak memory consumption during calibration. With this approach, calibration programs that previously required an 80GB A100 can now be executed using three 24GB RTX3090 GPUs for 7b models. We compare the memory usage of MBQ and our method using two different models, Qwen2-VL-7b \& LLaVA-onevision-7b, and three different calibration sample settings. As shown in the table \ref{tab:memory}, our method achieves substantially lower single-GPU peak memory usage compared to MBQ, and the 7B model can be calibrated normally with 128 calibration samples on three RTX3090 GPUs.

\begin{table}[h] 
\centering 
\footnotesize
\renewcommand{\arraystretch}{0.9}
\caption{Memory Usage Experiment During Calibration. The single GPU peak memory consumption was measured for models of 7b with varying numbers of calibration set samples. The experiments were conducted on three RTX3090 for 7b models. "OOM" indicates Out of Memory.}
\label{tab:memory}

\begin{tabular}{cccc}
\toprule
\multirow{2}{*}{Model} & \multirow{2}{*}{samples} & \multicolumn{2}{c}{peak memory(GB)} \\
\cmidrule{3-4}
& & MBQ & TLQ \\
\midrule
\multirow{3}{*}{Qwen2-VL-7b}
& 32    &   8.7   &  7.1   \\
& 64    &  16.4  &  7.3 \\
& 128   &\cellcolor{red!10}{ OOM }  &  10.9  \\
\midrule
\multirow{3}{*}{LLaVA-onevision-7b}
& 32    &  11.6  & 7.6   \\
& 64    &  \cellcolor{red!10}{OOM }  &  8.7  \\
& 128   & \cellcolor{red!10}{OOM }  &  18.8  \\
\bottomrule
\end{tabular}
\end{table}

\section{Conclusion}
This work rethinks PTQ for VLMs from the perspective of calibration alignment and identifies token-level misalignment as a key source of calibration bias. For this issue, we propose a \textbf{T}oken-level Importance-aware \textbf{L}ayer-wise \textbf{Q}uantization framework. Specifically, TLQ leverages gradient-guided token-level importance estimation to construct token-level calibration sets, enabling finer-grained calibration. In addition, we introduce a quantization-exposed, multi-GPU layer-wise calibration strategy that keeps calibration consistent with the true quantized inference path and distributes quantization-related computations across multi-GPU platforms, reducing reliance on large-memory GPUs. Extensive experiments 
across three model scales of two representative VLMs under two quantization settings 
show consistent performance improvements over strong PTQ baselines, highlighting the effectiveness of TLQ and its strong quantization stability.

\newpage
\section*{Impact Statement}
This paper presents work whose goal is to advance the field of Machine Learning. There are many potential societal consequences of our work, none which we feel must be specifically highlighted here.

\bibliography{example_paper}
\bibliographystyle{icml2026}

\newpage
\appendix
\onecolumn
\section{Supplementary Material}

\subsection{Additional experiments under the W8A8 quantization setting}

\begin{table*}[h]
\centering
\footnotesize
\renewcommand{\arraystretch}{0.9}
\caption{fp16 and W8A8 results of MBQ and our TLQ on LLaVA-onevision-7b and Qwen2-VL-7b. To ensure fair comparison and consistency with prior evaluations, we reuse the results of overlapping experimental settings reported in MBQ. Methods marked with * means the results of the first four benchmarks are reused.}
\label{tab:w8a8}
\resizebox{\textwidth}{!}{
\begin{tabular}{lccccccccc}
\toprule
Model & Bitwidth & Method & MMMU & OCRBench & VizWiz & TextVQA & ChartQA & SEEDBench2plus & Average($\uparrow$) \\
\midrule
\multirow{3}{*}{LLaVA-onevision-7b} & fp16 & - 
& 46.0 & 62.2 & 60.4 & 76.1 & 80.0 & 64.8 & 64.9 \\
\cmidrule{2-10}
&\multirow{2}{*}{W8A8}
&MBQ*  & 45.6 & 62.6 & 61.0 & 75.7 & 80.2 & 64.9 & 65.0 \\
&&TLQ & 46.2 & 61.7 & 60.5 & 76.1 & 79.8 & 65.0 & 64.9 \\

\midrule
\multirow{3}{*}{Qwen2-VL-7b} & fp16 & - 
& 50.6 & 80.7 & 68.3 & 82.0 & 81.4 & 69.4 & 72.1 \\
\cmidrule{2-10}
&\multirow{2}{*}{W8A8}
&MBQ*  & 50.1 & 80.7 & 68.3 & 81.8 & 81.3 & 69.1 & 71.9 \\
&&TLQ & 49.8 & 81.0 & 68.4 & 81.8 & 81.3 & 69.1 & 71.9 \\

\bottomrule
\end{tabular}
}
\end{table*}

For LLaVA-onevision-7b and Qwen2-VL-7b model, we also conduct experiments under the W8A8 quantization setting for MBQ and our TLQ. Results are reported in the table \ref{tab:w8a8}. Under this configuration, both our method and MBQ achieve performance close to the fp16 baseline. This indicates that high-bit quantization largely preserves model capacity, leaving limited room for performance differentiation among quantization strategies. Consequently, the advantages of our method become more pronounced under more aggressive low-bit quantization settings, where quantization errors play a more critical role.

\subsection{Gradient Visualization Analysis Before and After Token Selection}
\begin{figure}[h]
    \centerline{\includegraphics[width=\columnwidth]{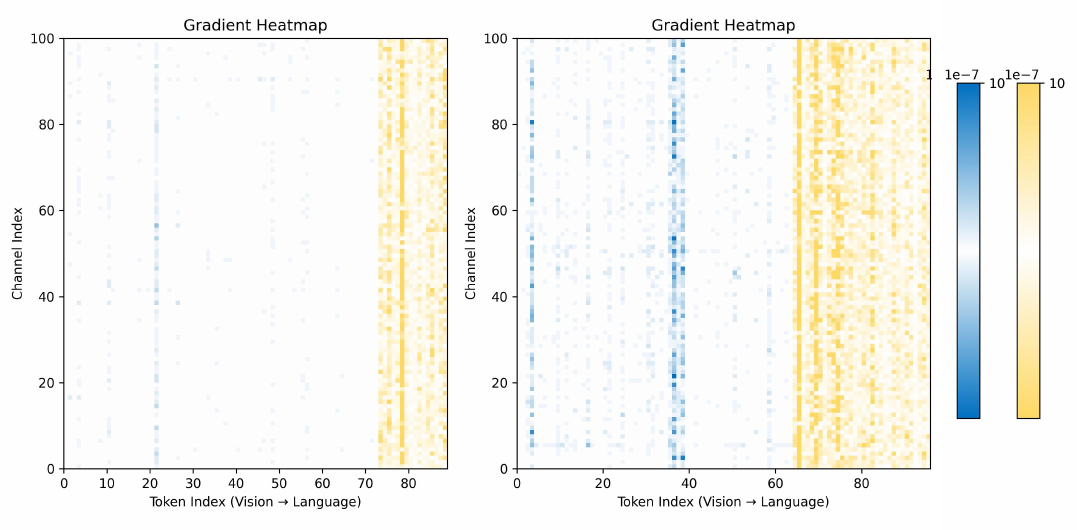}}
    \caption{Visualization of token gradients of 13th transformer block in the LLaVA-onevision-7b on COCO caption dataset before(left) and after(right) our gradient-guided token selection. The blue regions correspond to gradients of visual tokens, the yellow regions represent gradients of textual tokens, and the white areas indicate zero gradients.}
    \label{img：selection}
\end{figure}

We perform a visualization analysis of token gradients before and after token selection, as shown in the figure \ref{img：selection}.
The original gradient of a single sample has a shape of 1058 × 3584. For visualization, we randomly sample both channels and tokens while preserving the original ratio between visual and textual tokens, ensuring that each token remains sufficiently large for clear visualization.

It can be observed that the original model contains a large number of redundant visual tokens with zero gradients, and the proportion of visual tokens is significantly higher than that of textual tokens. Directly computing the scale using these activations may cause the scale to be biased by the redundant visual tokens, which in turn amplifies the quantization error of the model’s final output. By applying gradient-based token selection, the number of zero-gradient visual tokens is significantly reduced, increasing the relative proportion of textual tokens. As a result, only tokens with larger gradients contribute to the scale computation, thereby reducing the quantization error of the model’s final output.

\end{document}